# AVALIAÇÃO DA DOENÇA DE ALZHEIMER PELA ANÁLISE MULTIESPECTRAL DE IMAGENS DW-MR POR REDES RBF COMO ALTERNATIVA AOS MAPAS ADC

**Wellington P. dos Santos[1], Ricardo E. de Souza[2], Ascendino F. Dias e Silva[3] e Plínio B. dos Santos Filho[4]**

[1]Departamento de Sistemas Computacionais
Escola Politécnica de Pernambuco
Universidade de Pernambuco
50720-001, Recife, Pernambuco
e-mail: wps@dsc.upe.br

[2]Departamento de Física
Universidade Federal de Pernambuco
50670-901, Recife, Pernambuco
e-mail: res@df.ufpe.br

[3]Departamento de Eletrônica e Sistemas
Universidade Federal de Pernambuco
50740-530, Recife, Pernambuco
e-mail: afds@ufpe.br

[4]Department of Physics
North Carolina State University
Raleigh, North Carolina, EUA
e-mail: c2511@terra.com.br

**Resumo** - A doença de Alzheimer é a causa mais comum de demência, ainda difícil de diagnosticar com precisão sem o uso de técnicas invasivas, particularmente no começo da doença. Este trabalho aborda classificação e análise de imagens sintéticas multiespectrais compostas por volumes cerebrais de ressonância magnética (RM) ponderados em difusão para avaliação da área de fluido cérebro-espinal e sua correlação com o avanço da doença de Alzheimer. Um sistema de imagens de RM de 1,5 T foi utilizado para adquirir todas as imagens apresentadas. Os métodos de classificação são baseados em perceptrons multicamadas e classificadores de redes de função de base radial. Assume-se que as classes de interesse podem ser separadas por hiperquádricas. Uma rede polinomial de grau 2 é utilizada para classificar os volumes originais, gerando um volume verdade. Os resultados de classificação são utilizados para melhorar a análise usual pelo mapa de coeficientes de difusão aparentes.

**Palavras-chave** - doença de Alzheimer, imagens de ressonância magnética, análise multiespectral, classificadores baseados em redes neurais, redes de função de base radial

## 1. Introdução

A doença de Alzheimer é a maior causa de demência, tanto no grupo etário senil quanto no pré-senil, observando-se seu aumento gradual à medida em que o indivíduo envelhece [1,2]. As principais manifestações da doença compreendem o comprometimento cognitivo com gradual perda de memória, além de sintomas psicológicos, neurológicos e comportamentais que indicam declínio nas atividades da vida diária como um todo [3]. O mal de Alzheimer é caracterizado pela redução da matéria cinzenta e pelo aumento dos sulcos. A matéria cinzenta é responsável pela memória e sua redução explica a perda gradual de memória no indivíduo senil afetado por esta doença. Entretanto, também a matéria branca é afetada, apesar de ser desconhecida a relação entre a doença e a matéria branca [4-7].

A aquisição de imagens de ressonância magnética ponderadas em difusão possibilita a visualização da dilatação dos cornos temporais dos ventrículos laterais, além de realçar o aumento dos sulcos, que estão relacionados com o avanço da doença de Alzheimer [8]. Assim, a medição volumétrica de estruturas cerebrais é de grande importância para o diagnóstico e a avaliação do progresso de doenças como o mal de Alzheimer [1,9-11], em especial a medição das áreas ocupadas pelos sulcos e pelos ventrículos laterais, pois estas medidas permitem a adição de informação quantitativa à informação qualitativa expressa pelas imagens de ressonância magnética ponderadas em difusão [12].

A avaliação do progresso da doença de Alzheimer por meio da análise de imagens de ressonância magnética ponderadas em difusão é feita a partir da aquisição de no mínimo três imagens de cada fatia de interesse, onde cada uma dessas imagens é adquirida usando a seqüência eco de spin Stejskal-Tanner, com expoentes de difusão diferentes, sendo um deles igual a 0

s/mm$^2$. Ou seja, uma das três imagens é uma imagem ponderada em $T_2$ [5,8]. Utilizando essas três imagens assim adquiridas, é montada uma quarta imagem, o mapa ADC, ou mapa de coeficientes de difusão aparente (*Apparent Diffusion Coefficient Map*), onde cada *pixel* corresponde ao coeficiente de difusão aparente do *voxel* correspondente. Quanto maior o brilho do *pixel* no mapa ADC, maior o coeficiente de difusão aparente [8].

Este trabalho propõe uma nova abordagem para avaliação do progresso da doença de Alzheimer: uma vez que o mapa ADC usualmente apresenta *pixels* com intensidades não nulas em regiões não ocupadas pela amostra, alguma incerteza também pode ser levantada quanto aos coeficientes de difusão do interior da amostra, sendo sensível à presença de ruídos nas imagens [8]. Assim, neste estudo de caso, as imagens são utilizadas na composição de uma única imagem multiespectral, onde cada uma das imagens ponderadas em difusão se comporta como uma banda espectral de uma imagem multiespectral sintética. A imagem multiespectral sintética assim montada é então classificada utilizando dois métodos: as redes neurais perceptrons multicamadas (MLP) e as redes de função de base radial.

## 2. Materiais e Métodos

### 2.1. Imagens Ponderadas em Difusão e Mapas ADC

As imagens de ressonância magnética (RM) ponderadas em difusão foram obtidas a partir do banco de imagens clínicas do Laboratório de Imagens de Ressonância Magnética do Departamento de Física da Universidade Federal de Pernambuco. O banco é composto de imagens clínicas reais obtidas por um tomógrafo clínico de RM de 1,5 T.

Neste trabalho foram utilizadas 80 imagens de ressonância magnética axiais ponderadas em difusão, ou seja, 4 volumes cerebrais de 20 fatias axiais cada, sendo um desses volumes composto por mapas ADC, correspondentes a um paciente do sexo masculino, de 70 anos de idade, portador da doença de Alzheimer. Para realizar o treinamento foi escolhida a fatia 13 de cada amostra volumétrica de 20 fatias, dado que essa fatia mostra os cornos temporais dos ventrículos laterais, o que permite uma avaliação mais clara por parte do especialista e facilita a correlação entre os dados gerados pela ferramenta computacional e o conhecimento *a priori* do especialista. Além disso, a fatia 13 apresenta artefatos fora da região cranial em quantidade considerável.

Podem-se considerar imagens como funções matemáticas, onde seu domínio é uma região do plano dos inteiros, a *grade*, e seu contradomínio é o conjunto dos valores possíveis de serem assumidos pelos *pixels* correspondentes a cada posição da grade.

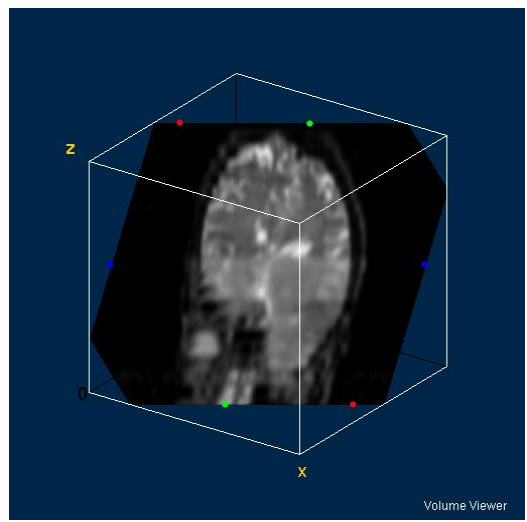

Figura 1. Volume ponderado em difusão com expoente de difusão 0 s/mm$^2$

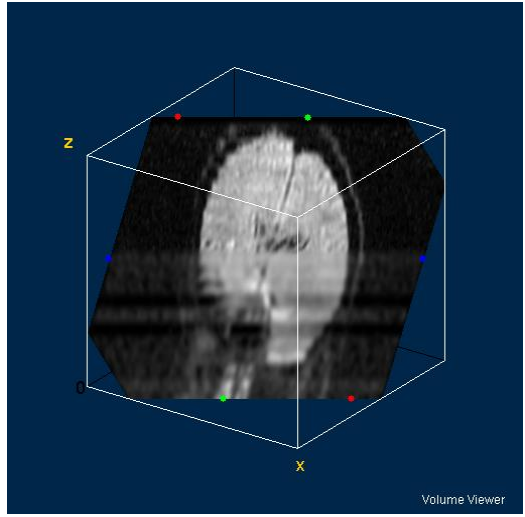

Figura 2. Volume ponderado em difusão com expoente de difusão 500 s/mm$^2$

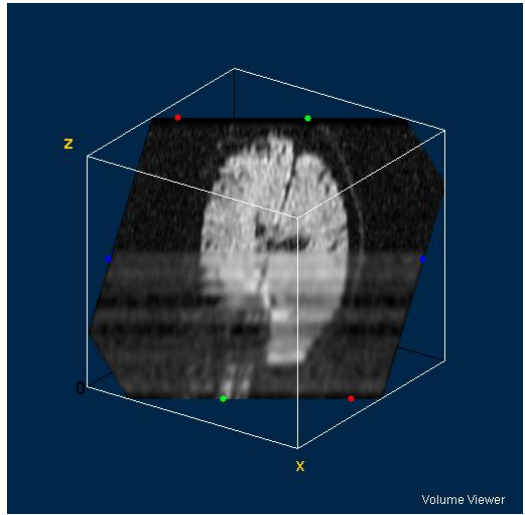

Figura 3. Volume ponderado em difusão com expoente de difusão 1000 s/mm$^2$

Seja $f_i : S \to W$ o conjunto das imagens de RM ponderadas em difusão, onde $1 \leq i \leq 3$, $S \subseteq \mathbf{Z}^2$ é a grade da imagem $f_i$, onde $W \subseteq \mathbf{R}$ é seu contradomínio. Cada imagem multiespectral sintética $f : S \to W^3$ integrante do volume multiespectral a ser classificado é composta pelas 3 imagens de RM ponderadas em difusão obtidas dos volumes mostrados nas figuras 1, 2 e 3, sendo dada por:

$$f(\mathbf{u}) = (f_1(\mathbf{u}), f_2(\mathbf{u}), f_3(\mathbf{u}))^T, \qquad (1)$$

onde $\mathbf{u} \in S$ é a posição do *pixel* na imagem $f$, e $f_1$, $f_2$ e $f_3$ são as imagens de RM de difusão. Considerando que cada *pixel* $f_i(\mathbf{u})$ é aproximadamente proporcional ao sinal do *voxel* correspondente como segue [13-19]:

$$f_i(\mathbf{u}) = K\rho(\mathbf{u})e^{-T_E/T_2(\mathbf{u})}e^{-b_i D_i(\mathbf{u})}, \qquad (2)$$

onde $D_i(\mathbf{u})$ é o coeficiente de difusão de *spin* associado ao *voxel* da posição $\mathbf{u}$, medido após o $i$-ésimo experimento; $\rho(\mathbf{u})$ é a densidade de *spins* nucleares no referido *voxel*; $K$ é uma constante de proporcionalidade; $T_2(\mathbf{u})$ é o tempo de relaxação transversal no *voxel*; $T_E$ é o tempo de eco e $b_i$ é o expoente de difusão, dado por [8,20,21]:

$$b_i = \frac{1}{3}\gamma^2 G_i^2 T_E^3, \qquad (3)$$

onde $\gamma$ é a constante giromagnética e $G_i$ é o gradiente aplicado no $i$-ésimo experimento. As figuras 1, 2 e 3 mostram os volumes de imagens ponderadas em difusão com expoentes de difusão 0 s/mm², 500 s/mm² e 1000 s/mm², respectivamente.

A análise de imagens de RM ponderadas em difusão é usualmente realizada através dos mapas ADC, $f_{ADC}: S \to W$, calculados como segue [22]:

$$f_{ADC}(\mathbf{u}) = \frac{C}{b_2}\ln\left(\frac{f_1(\mathbf{u})}{f_2(\mathbf{u})}\right) + \frac{C}{b_3}\ln\left(\frac{f_1(\mathbf{u})}{f_3(\mathbf{u})}\right), \qquad (4)$$

onde $C$ é uma constante de proporcionalidade.

Considerando $n$ experimentos, pode-se generalizar a equação 4 como segue:

$$f_{ADC}(\mathbf{u}) = \sum_{i=2}^{n}\frac{C}{b_i}\ln\left(\frac{f_1(\mathbf{u})}{f_i(\mathbf{u})}\right). \qquad (5)$$

Portanto, o mapa ADC é dado por:

$$f_{ADC}(\mathbf{u}) C \overline{D}(\mathbf{u}), \qquad (6)$$

onde $\overline{D}(\mathbf{u})$ é a média amostral das medidas do coeficiente de difusão $D(\mathbf{u})$ [15,23,24].

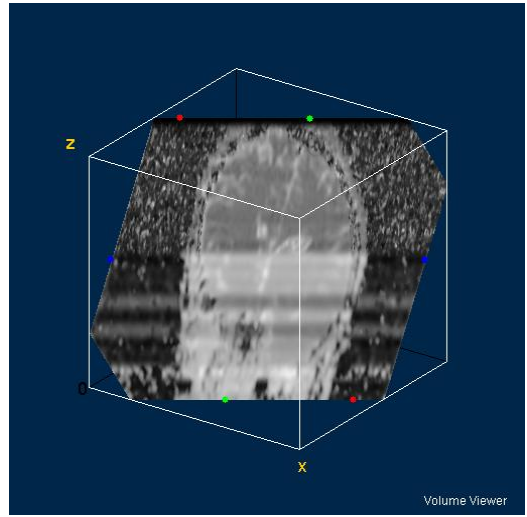

Figura 4. Volume composto de mapas ADC calculados a partir das imagens de difusão dos 3 volumes anteriores

Portanto, os *pixels* do mapa ADC são proporcionais aos coeficientes de difusão nos *voxels* correspondentes. Entretanto, como as imagens são adquiridas em diferentes momentos, deve-se considerar a ocorrência de ruído em todos os experimentos. Na figura 4 pode-se notar diversos artefatos devidos à presença de ruído. Nas regiões da imagem onde a relação sinal-ruído é pobre (por exemplo, $s/n \cong 1$), o mapa ADC produz artefatos como conseqüência do cálculo do logaritmo (ver equações 4 e 5). Esses fatores levam à seguinte conclusão: os *pixels* do mapa ADC não necessariamente correspondem aos coeficientes de difusão, uma vez que diversos *pixels* indicam altas taxas de difusão em *voxels* onde simplesmente não há amostra ou em áreas muito sólidas, como a caixa cranial, por exemplo, como pode ser notado na figura 4. Essa é a razão pela qual esses mapas indicam coeficientes de difusão *aparentes*, e não *reais*.

Neste trabalho propõe-se uma alternativa à análise do mapa ADC: a análise multiespectral do volume composto pelas imagens $f : S \to W^3$ utilizando métodos baseados em redes neurais.

## 2.2. Análise Multiespectral utilizando Redes Neurais

Seja o universo de classes de interesse $\Omega = \{C_1, C_2, C_3\}$, $C_1$ representa os coeficientes de difusão associados ao fluido cérebro-espinal; $C_2$, matéria branca e cinzenta, uma vez que não podem ser distinguidas por meio de imagens de difusão, pois seus coeficientes de difusão são muito próximos; $C_3$ corresponde ao fundo da imagem.

Para a análise multiespectral utilizando redes neurais, as entradas são associadas ao vetor $\mathbf{x} = (x_1, x_2, x_3)^T$, onde $x_i = f_i(\mathbf{u})$, para $1 \leq i \leq 3$. As saídas da rede representam as classes de interesse e estão associadas ao vetor $\mathbf{y} = (y_1, y_2, y_3)^T$, onde cada saída corresponde à classe de mesmo índice. O critério de decisão é baseado no critério de Bayes: a maior saída indica a classes mais provável [25,26]. O conjunto de treinamento é construído utilizando conhecimento especialista durante a seleção das regiões de interesse [27].

O volume de imagens multiespectrais sintéticas foi classificado usando os seguintes métodos:

1) Perceptron multicamadas (MLP): Taxa de aprendizado inicial $\eta_0 = 0,2$, erro de treinamento $\varepsilon = 0,05$, máximo de 1000 iterações, 3 entradas, 3 saídas, 2 camadas, 60 neurônios na camada 1 [27];

2) Rede de função de base radial (RBF): 3 entradas, 2 camadas; camada 1: mapa de k-médias com 18 neurônios na camada 1, taxa de aprendizado inicial $\eta_0 = 0,1$, máximo de 200 iterações; camada 2: 3 saídas, máximo de 200 iterações, taxa de aprendizado inicial $\eta_0' = 0,1$ [27].

O perceptron multicamadas foi escolhido para se avaliar o desempenho da classificação multiespectral baseada em redes neurais clássicas de duas camadas. O número de entradas e saídas corresponde ao número de bandas e de classes de interesse, respectivamente. O erro de treinamento foi escolhido considerando o máximo ruído estimado em imagens ponderadas em difusão. O número de neurônios na camada 1 e a taxa de aprendizado foram determinados empiricamente.

A rede de função de base radial foi escolhida para avaliar o desempenho da classificação multiespectral baseada em uma estratégia orientada a problemas locais. O número de entradas e saídas corresponde ao número de bandas e de classes de interesse, respectivamente. A taxa de aprendizado foi determinada empiricamente.

Para implementar os métodos utilizados e reconstruir os volumes, foi desenvolvida uma ferramenta computacional que recebeu o nome de AnImed, construída utilizando a linguagem de programação orientada a objeto Object Pascal no ambiente de desenvolvimento Delphi 5. A visualização dos volumes foi efetuada por meio da ferramenta baseada em Java, ImageJ, desenvolvida pelo NIH (*National Institute of Health*, EUA) e do *plugin* VolumeViewer.

## 2.3. Análise do Mapa ADC usando Fuzzy C-Médias

Para comparar os métodos propostos e o mapa ADC é preciso extrair informação quantitativa e qualitativa do mapa ADC. Isso é possível aplicando uma classificação monoespectral não-supervisionada ao mapa ADC utilizando um método baseado em agrupamento [28,29]. Foi escolhido então o *fuzzy* c-médias (ADC-CM) com 3 entradas, 3 saídas, máximo de 200 iterações e taxa de aprendizado inicial $\eta_0 = 0,1$.

As 3 entradas são associadas ao vetor $\mathbf{x} = (x_1, x_2, x_3)^T$, onde $x_i = f_{ADC}(\mathbf{u})$, para $1 \leq i \leq 3$. As saídas da rede representam as classes de interesse e estão associadas ao vetor $\mathbf{y} = (y_1, y_2, y_3)^T$, onde cada saída corresponde à classe de mesmo índice. O critério de decisão empregado é o mesmo da subseção anterior: a maior saída indica a classe mais provável [25,26]. O conjunto de treinamento é composto por *pixels* da imagem $f_{ADC}$ e construído usando conhecimento especialista na seleção das regiões de interesse [27].

## 3. Resultados

Para avaliar objetivamente os resultados da classificação volumétrica, foram utilizados três métodos: o *índice* $\kappa$, a *acurácia global* e a *matriz de confusão*. A avaliação subjetiva foi realizada utilizando conhecimento especialista. As classes fundo da imagem ($C_3$), matéria branca e cinzenta ($C_2$), e líquido cérebro-espinal ($C_1$) foram associadas às seguintes cores: preto, cinza e branco, respectivamente. A figura 5 mostra o conjunto de treinamento montado sobre a fatia 13 do volume de mapas ADC, enquanto a figura 6 mostra o volume verdade.

Para a tarefa de classificação, assume-se que as classes de interesse são separáveis por hiperquádricas. Portanto, escolhe-se uma rede polinomial para classificar o volume multiespectral original e gerar um volume verdade. O grau do polinômio foi empiricamente determinado pelo seu incremento gradual até que não houvesse diferenças significativas entre a classificação presente e a classificação imediatamente anterior.

A rede polinomial é uma rede de duas camadas: a primeira é uma rede multiplicativa que gera todos os termos do polinômio de grau 2 a partir das 3 entradas; a segunda camada consiste em um perceptron de camada única com taxa de aprendizado inicial $\eta_0 = 0,1$ e erro de treinamento $\varepsilon = 0,05$, máximo de 200 iterações de treinamento, responsável pelo cálculo dos coeficientes do polinômio que modela as funções discriminantes de cada classe [25,26]. A rede polinomial é um aproximador polinomial. A taxa de aprendizado e o erro de treinamento foram determinados empiricamente.

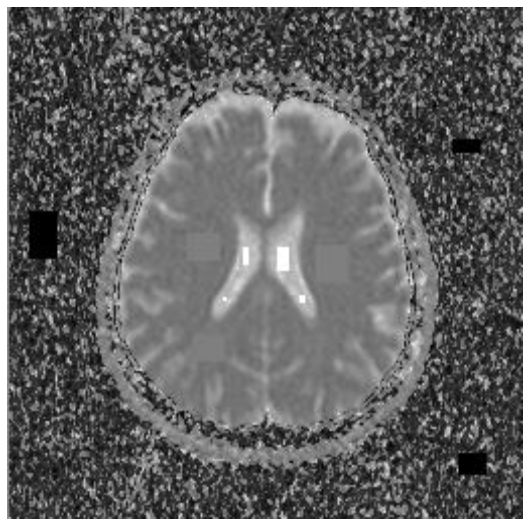

Figura 5. Conjunto de treinamento

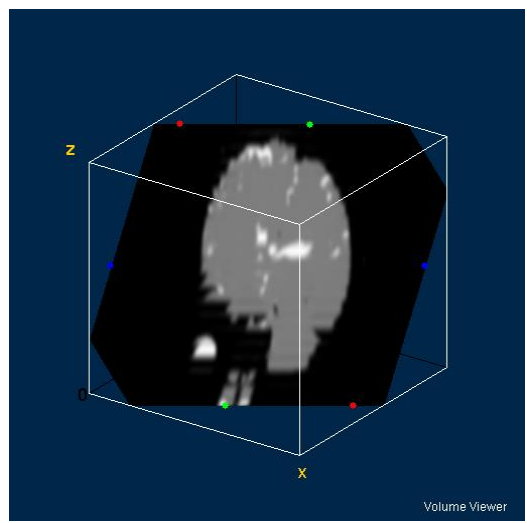

Figura 6. Volume verdade gerado pela classificação usando a rede polinomial

A *matriz de confusão* para o universo de classes de interesse $\Omega$ é uma matriz $m \times m$, $T = [t_{i,j}]_{m \times m}$, onde cada elemento $t_{i,j}$ representa o número de objetos pertencentes à classe $C_j$ classificados como $C_i$, onde $\Omega = \{C_1, C_2, \ldots, C_m\}$ [25,30].

A *acurácia global* $\phi$ é a razão entre o número de objetos corretamente classificados e o total de objetos, definida como segue [25,30]:

$$\phi = \rho_v = \frac{\sum_{i=1}^{m} t_{i,i}}{\sum_{i=1}^{m} \sum_{j=1}^{m} t_{i,j}}. \tag{7}$$

O *índice* $\kappa$ é um índice de correlação estatística definido como segue [25]:

$$\kappa = \frac{\rho_v - \rho_z}{1 - \rho_z}, \tag{8}$$

onde:

$$\rho_z = \frac{\sum_{i=1}^{m} \left(\sum_{j=1}^{m} t_{i,j}\right)\left(\sum_{j=1}^{m} t_{j,i}\right)}{\sum_{i=1}^{m} \sum_{j=1}^{m} t_{i,j}^{2}}. \tag{9}$$

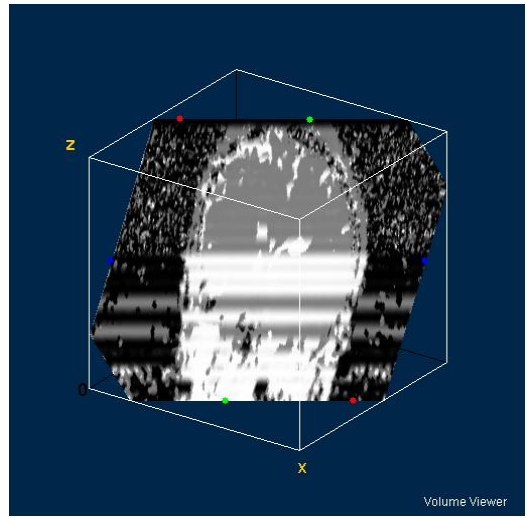

Figura 7. Classificação pelo método *fuzzy* c-médias usando o mapa ADC

A figura 7 mostra o resultado da classificação do mapa ADC usando o método *fuzzy* c-médias. As figuras 8 e 9 mostram os resultados da classificação do volume de imagens multiespectrais sintéticas composto pelos volumes das figuras 1, 2 e 3 usando os métodos MLP e RBF, respectivamente. A tabela 1 apresenta o índice $\kappa$ e a acurácia global $\phi$, enquanto a tabela 2 mostra os volumes percentuais $V_1$, $V_2$ e $V_3$ ocupados pelas classes de interesse $C_1$, $C_2$ e $C_3$, nesta ordem, bem como a razão entre o volume de líquido cérebro-espinal e o volume total de matéria branca e cinzenta, denominada simplesmente de razão fluido-matéria, expressa por $V_1 / V_2$.

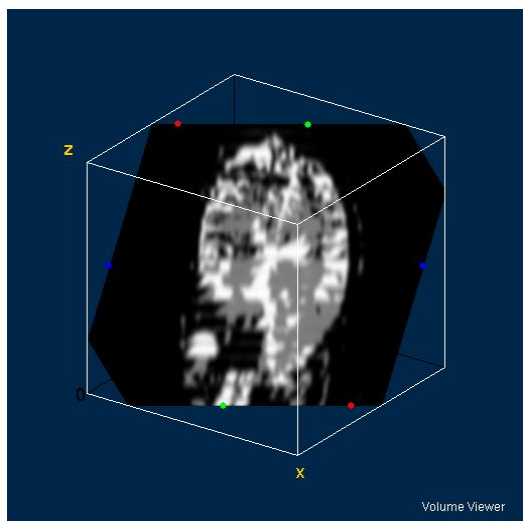

Figura 8. Classificação pelo perceptron multicamadas

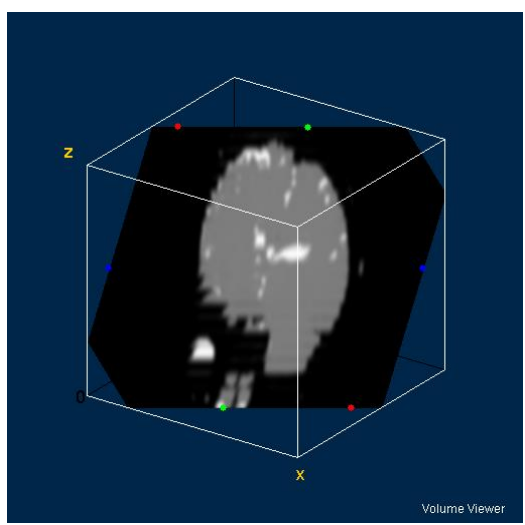

Figura 9. Classificação pela rede de função de base radial

Tabela 1: Acurácia global $\phi$ (%) e índice $\kappa$ para os métodos de classificação

|  | MLP | BRF | ADC-CM |
|---|---|---|---|
| $\phi$ (%) | 88,5420 | 99,3587 | 58,1154 |
| $\kappa$ | 0,6081 | 0,9681 | 0,2495 |

Tabela 2: Volumes percentuais e razão fluido-matéria para os métodos de classificação

|  | MLP | RBF | ADC-CM | PO |
|---|---|---|---|---|
| $V_1$ (%) | 7,607 | 1,612 | 18,743 | 1,697 |
| $V_2$ (%) | 11,546 | 17,187 | 34,354 | 17,010 |
| $V_3$ (%) | 80,847 | 81,201 | 46,903 | 81,293 |
| $V_1/V_2$ | 0,659 | 0,094 | 0,546 | 0,100 |

## 4. Discussão e Conclusões

Os resultados da classificação multiespectral forneceram uma melhor avaliação, tanto quantitativa quanto qualitativa, dos volumes adquiridos, tornando possível a medição dos volumes de interesse, em especial a razão entre o volume ocupado pelo líquido cérebro-espinal e o volume ocupado pelas matérias branca e cinzenta, correlacionando essa razão com o avanço da doença de Alzheimer.

Da tabela 1 pode-se notar que a abordagem multiespectral, com índice $\kappa$ de 0,6081 e 0,9681 para o perceptron multicamadas e para a rede de função radial, respectivamente, é superior à análise do mapa ADC pelo método *fuzzy* c-médias, com índice $\kappa$ de 0,2495.

Esses resultados são confirmados quando se observa os volumes classificados das figuras 8 e 9 e se compara com o volume classificado ADC da figura 7, onde se pode notar diversas áreas fora da amostra e na caixa cranial marcadas como líquido cérebro-espinal ou matéria branca ou cinzenta.

A partir dos resultados obtidos fica claro que o perceptron multicamadas superestimou o volume ocupado pelo líquido cérebro-espinal (ver figura 8). Quanto este resultado é comparado com o volume com expoente de difusão 0 (ver figura 1), pode-se perceber que os ventrículos esquerdo e direito são separados. Além disso, os sulcos também foram superestimados, o que poderia levar um especialista a talvez considerar esse caso de Alzheimer como mais avançado do que realmente é.

A superestimação dos volumes também pode ser percebida nos resultados na tabela 2, onde a razão fluido-matéria ($V_1/V_2$) é de 0,659 para o perceptron multicamadas. Da mesma ordem, portanto, do resultado obtido pela classificação do mapa ADC pelo método *fuzzy* c-médias, com $V_1/V_2 = 0,546$, quase 6 vezes a taxa obtida pela rede de função de base radial, com $V_1/V_2 = 0,094$, muito próxima, portanto, daquela obtida pela rede polinomial, $V_1/V_2 = 0,100$. Conseqüentemente, a rede RBF pode ser considerada uma boa estimadora da razão fluido-matéria, enquanto a rede MLP pode ser descartada.

A classificação multiespectral de imagens de RM ponderadas em difusão fornece uma boa alternativa à análise do mapa ADC, consistindo em uma ferramenta matemática bastante útil para analisar tanto quantitativa quanto qualitativamente o progresso da doença de Alzheimer pelo especialista.

As imagens utilizadas neste trabalho se limitam apenas a um único paciente porque ainda não existem bancos de imagens de ressonância magnética ponderadas em difusão (*Diffusion-Weighted Magnetic Resonance Images*, DW-MR) disponíveis para uso da comunidade científica. Isto se pode dever ao fato de, em geral, não se utilizar com freqüência imagens ponderadas em difusão para avaliações anatômicas, como é uma das propostas deste artigo, além de ser um tanto difícil gerar imagens DW-MR sintéticas relacionadas com determinadas patologias, dada a relação entre imagens DW-MR e a dinâmica dos fluidos envolvidos na amostra, no caso deste artigo, do líquor.

No caso particular da doença de Alzheimer, a geração de um banco de imagens DW-MR se torna mais difícil ainda, dada a complexidade da dinâmica envolvida na doença de Alzheimer, dinâmica esta ainda não plenamente conhecida pela comunidade científica. Além disso, até o momento ainda é muito difícil conseguir voluntários que possam fornecer mais imagens DW-MR de casos de Alzheimer. Um trabalho futuro poderia ser a construção de um banco de imagens sintéticas para simulação de vários casos patológicos, com imagens geradas a partir da obtenção de imagens de voluntários.

## 5. Agradecimentos

## 7. Referências Bibliográficas